\documentclass[journal]{IEEEtai}

\usepackage[colorlinks,urlcolor=blue,linkcolor=blue,citecolor=blue]{hyperref}

\usepackage{color,array}

\usepackage{graphicx}
\DeclareGraphicsExtensions{.pdf,.png}
\usepackage{amsmath} 
\usepackage{amssymb}  
\usepackage{tabularx,booktabs,multirow}
\usepackage{array}
\usepackage{makecell}
\usepackage{tablefootnote}
\usepackage{multicol}
\usepackage{booktabs}
\usepackage{threeparttable}
\usepackage{url}
\usepackage[ruled,vlined]{algorithm2e}

\begin{document}

\title{\LARGE \bf Vector Detection Network: An Application Study on Robots Reading Analog Meters in the Wild}

\author{Zhipeng Dong, Yi Gao, Yunhui Yan, and Fei Chen, \IEEEmembership{Senior Member, IEEE}
\thanks{This work is supported by the National Natural Science Foundation of China (51805078, 51374063) and the National Key Research and Development Program of China (2017YFB0304200). \textit{(Corresponding author: Fei Chen.)} }
\thanks{Z. Dong, Y. Gao, and Y. Yan are with the School of Mechanical Engineering and Automation, Northeastern University, Shenyang 110819, China (e-mail: zhipengdong@stumail.neu.edu.cn; yigaowork@163.com; yanyh@mail.neu.edu.cn).}
\thanks{F. Chen is with Smart Manipulation Robots Laboratory, Department of Mechanical and Automation Engineering, T-Stone Robotics Institute, The Chinese University of Hong Kong, Hong Kong (e-mail: f.chen@ieee.org).}
}


\maketitle

\begin{abstract}
Analog meters equipped with one or multiple pointers are wildly utilized to monitor vital devices' status in industrial sites for safety concerns. Reading these legacy meters {\bi autonomously} remains an open problem since estimating pointer origin and direction under imaging damping factors imposed in the wild could be challenging. Nevertheless, high accuracy, flexibility, and real-time performance are demanded. In this work, we propose the Vector Detection Network (VDN) to detect analog meters' pointers given their images, eliminating the barriers for autonomously reading such meters using intelligent agents like robots. We tackled the pointer as a two-dimensional vector, whose initial point coincides with the tip, and the direction is along tail-to-tip. The network estimates a confidence map, wherein the peak pixels are treated as vectors' initial points, along with a two-layer scalar map, whose pixel values at each peak form the scalar components in the directions of the coordinate axes. We established the Pointer-10K dataset composing of real-world analog meter images to evaluate our approach due to no similar dataset is available for now. Experiments on the dataset demonstrated that our methods generalize well to various meters, robust to harsh imaging factors, and run in real-time.
\end{abstract}

\begin{IEEEImpStatement}
From the automatic meter recognition perspective, this work proposes an accurate, flexible, and ready-to-use pointer detection algorithm that contributes to dealing with legacy pointer-type meters, which still serve in industrial sites to date and require extensive human inspection. Adopting this technique to automatic agents could save human labor and promote the effectiveness and efficiency of getting data. It also inspires particular object detection problems, wherein the object could be considered a vector whose position is concerned, and its orientation also matters. The proposed Vector Detection Network is agnostic to the semantic meaning of the vectors it detects. Thus it could also be used to detect objects of this kind, such as the branch of plants, helping the agriculture robots do accurate trimming.
\end{IEEEImpStatement}


\section{Introduction}

\IEEEPARstart{T}{he} inspection and surveillance in complex industrial sites, such as substations \cite{Allan2014}, nuclear power plants \cite{DiCastro2018}, and offshore platforms \cite{Bellicoso2018} play an essential role in terms of diagnosing device status, discovering abnormal, and preventing potential hazards. Such work has relied heavily on human labor, which makes no guarantee of data collection accuracy and puts tremendous stress on essential workers and data management. To address this issue, using autonomous agents to replace human could be an efficient alternative. Among all kinds of agents, movable patrol robots have been studied for decades \cite{Allan2014, Li2015, Tsitsimpelis2019} because they could travel in unstructured terrains while perceiving, processing, and transferring data with decent accuracy 24 hours a day, ensuring that the equipment's vital information is interpreted correctly.

\begin{figure}[t]
\centering
\includegraphics[width=\columnwidth]{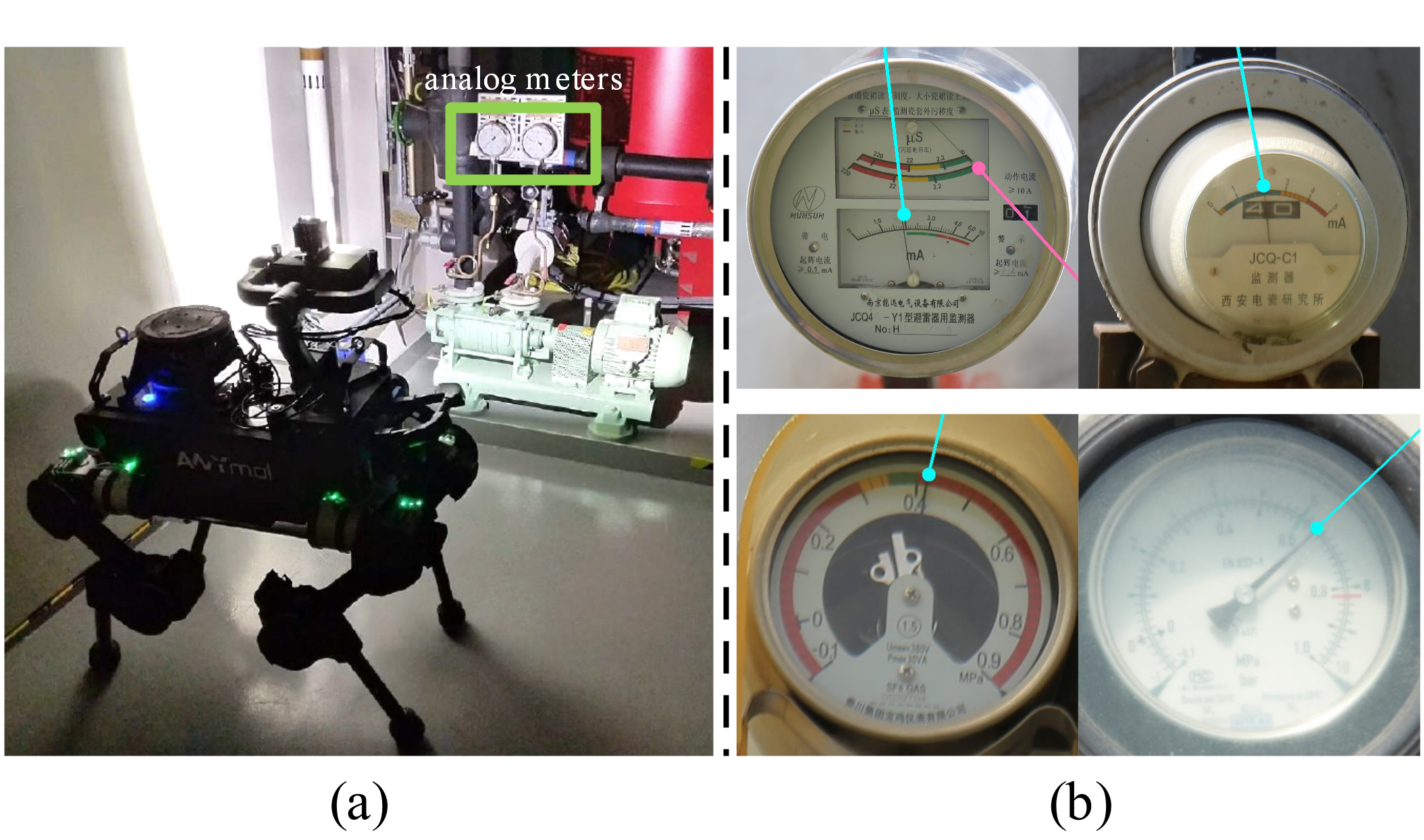}
\caption{(a) Illustration of the concept that a patrol robot inspects in the industrial site via reading pointer-type analog meters \cite{Gehring2019}. (b) The proposed neural network aids the reading procedure by detecting pointers in the captured image as vectors (depicted in colored rays). Despite the divergent characterizes of these meters captured in the wild, the model could deliver accurate results without fine-tuning.}
\label{fig:1}
\end{figure}

Although being beneficial, the adoption of automatic inspection is imperfect or even impossible if performing specific tasks could not avoid humans in the loop. In practice, a common category of jobs for patrol robots is to reveal vital equipment information via reading attached analog meters as shown in Fig. \ref{fig:1}(a), including manometers, ammeters, thermometer, and other kinds of meters with one or multiple pointers \cite{Gehring2019}. Due to concerns on cost, durability, and redundancy, these legacy meters tailored for manual recognition are unlikely to be replaced soon by intelligent instruments \cite{Singh2019}. Hence, it is essential to endow the patrol robots and its data processing system to read off them just like what humans do with their eyes and brains, which naturally leads to a vision-based scheme.

Such a scheme involves four image processing steps as detailed in Sec. \ref{Sec:3-A}, namely meter detection, pointer detection, template matching, and reading calculation. The meter detection issue has been investigated with methodologies ranging from circle Hough-transform (CHT) \cite{Wang2018}, histogram oriented gradients (HOG) \cite{Hutter2018}, perspective transform coupled with the sum of absolute differences (SAD) \cite{Cheng2020}, to learning-based methods \cite{Yang2018, Liu2020}. Moreover, since it is in the spectrum of object detection burgeoning in recent years, now an off-the-shelf technique such as Detectron2 \cite{Wu2019} can deliver acceptable detections with proper training. Meanwhile, the third and the fourth steps involving fundamental perspective transformation and geometric calculation could be tackled with well-established tools such as OpenCV \cite{Bradski2000}.

Unlike other steps, the pointer detection step is highlighted in this work because of its difficulty. This step takes images captured in the wild as input; thus, it must endure both intrinsic and extrinsic damping factors of the imaging. From the first perspective, the imaging of thin pointers captured by the onboard camera of the robot patrolling the industrial site tends to blur or vanish due to low image resolution, long-range zooming, and indecent exposure (incident to the high contrast between the dark foreground and the bright background). These drawbacks could be overcome to a extend by upgrading the camera, finding a better observation spot, or adjusting the imaging algorithm. However, that would raise the budget and impose more implementation constraints, yet could not substantially solve it. Second, the extrinsic factors presented in the environment also contribute to imaging quality negatively. The device's vibration could blur the pointer, the reflective glass covering the dial plate could be contaminated by dust or oil or too reflective under poor lighting conditions. Nevertheless, with all these adverse factors, the inspection algorithm must achieve an error rate lower than $\pm5\%$ according to first-line inspection force's requirements, avoids fine-tuning for each kind of meter, and better runs in real-time. The inadequate imaging quality and high demand for usability have put conventional techniques into a struggle, as detailed in Sec. \ref{Sec:2}. 

In this work, we propose a learning-based approach that detects the pointer as a two-dimensional Euclidean vector, which starts from the pinpoint and points along the tail-to-tip direction as illustrated in Fig. \ref{fig:1}(b). We were partially inspired by keypoint detection networks applied to human pose estimation \cite{Cao2017, Xiao2018}, which as our work, deal with real scenarios in the wild. Our architecture inherits the capabilities of these networks for handling the damping factors of imaging. However, distinct from these works, we detect the keypoints and simultaneously estimate their contextual directions in the image. Thus, the reading pipeline could be seamlessly aligned without bells and whistles to ascertain the pointers' orientation. To our knowledge, we are the first to apply such a technique in the pointer detection domain. The proposed Vector Detection Network (VDN) trained with tailored datasets (detailed in Sec. \ref{Sec:4}) could detect multiple instances in real-time and is free from hand-designed feature extraction procedures such as binarization, morphological operation, skeletonizing, or edge detection. It avoids extensive parameter adjustment for each type of meter on-site, which generalizes well to challenging circumstances and is ready to be used with patrol robots.

\section{Background}\label{Sec:2}

\begin{figure*}[t]
\centering
\vspace{9pt}
\includegraphics[width=7.16in]{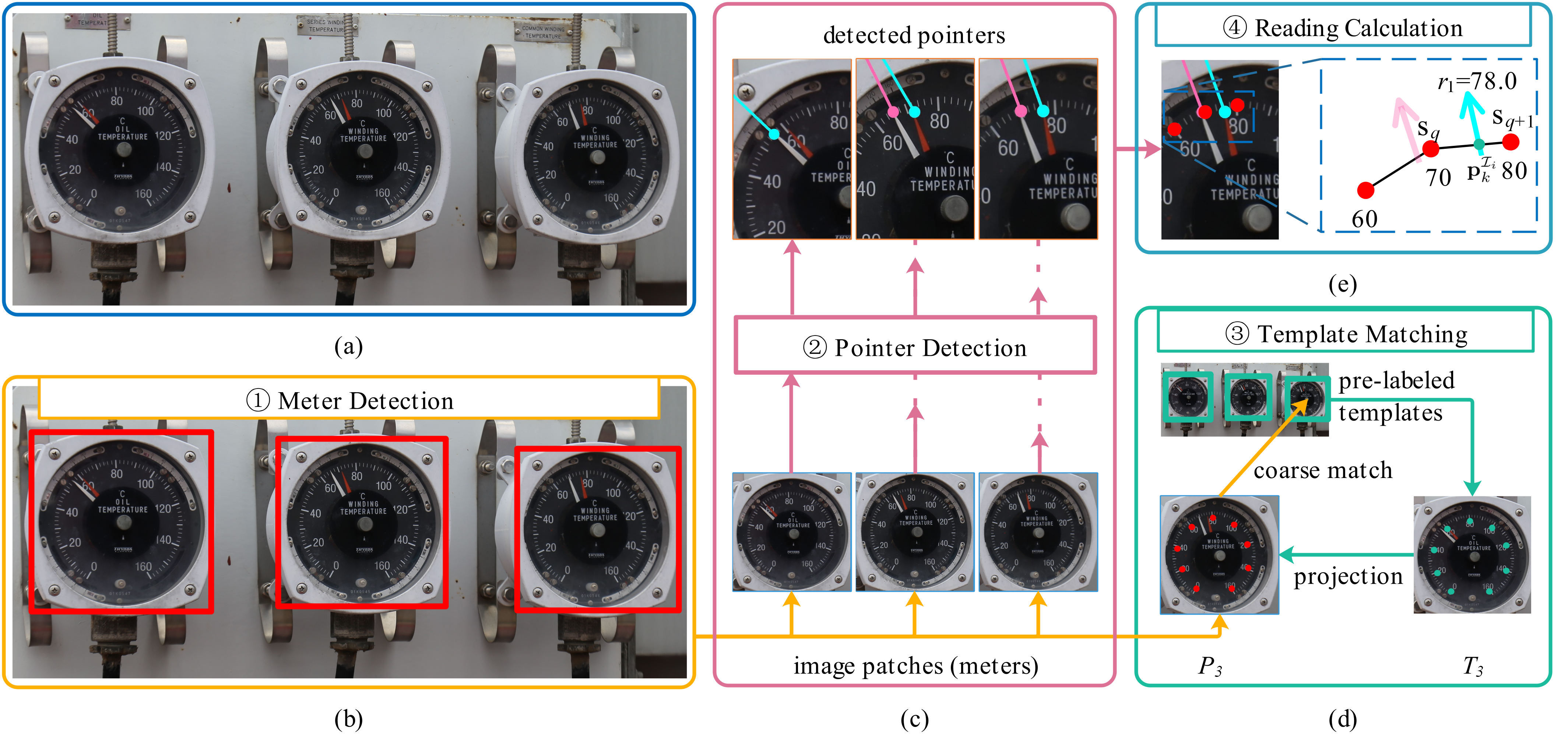}
\caption{The pipeline of reading analog meters in industrial sites with robots. Note that this pipeline does not include image preprocessing nor touch the details of each module. We would like to show that these modules are self-contained and could be implemented and improved independently. (a) Original image captured by the robot. (b) Results of the meter detection module with RCNN backend. (c) The pointer detection module processes the input image patches and outputs detected pointers for each patch. Note that coincide pointers were detected as one but treated separately. (d) The module projected the labeled physical scale points (green dots) on dial plates to those patches (red dots) by matching the template prepared beforehand with the image patches. (e) Given scale points positions and their values, the module calculated pointers' readings with \eqref{eq:2}.}
\label{fig:2}
\end{figure*}

Automatic meter recognition (AMR) is a broad domain dealing with meters with no direct data I/O with the outside world but relying on visual interfaces, such as pointers or light-emitting diode displays, to reveal the information. This section focuses on recognizing pointer-type meters, which on their own include a wide range of species and variants. Since this work aims to address the bottleneck, i.e., pointer detection, we first go through conventional automatic recognition pipelines and then narrow down to the pointer detection module, where this work made a distinct advance over traditional methods in terms of accuracy, robustness, and flexibility.

A significant part of the literature in this domain leverage classical image processing techniques to formalize the recognition pipeline, which often starts by detecting the \textit{primitives} of the meter in the image. Such primitives include pointers, scales, and lettering elements \cite{Sablatnig1994}. From the geometric relations and constraints between the pointer and the scales, one could numerically read the meter, and by considering the labels, one could infer the actual semantic reading. Such a scheme closely follows human intuition and is straightforward to implement because it separates the pipeline into isolated modules, which could be addressed one by one. 

The pointer detection module deals with pointers rotating or sliding across the meter plane, making it harder to build than the static scales or labels' detection modules, which could be interpreted with prior knowledge or template matching. Using prior knowledge, such as the pointer's color as the cue, could be too vulnerable to generalize to real scenarios. Using an exhaustive method, like matching the image with templates encoding different pointer positions, could be inefficient and inaccurate. As a result, the mainstream of pointer detection relies on the geometric factor, treating a pointer as a line segment. Hence, the line detection method could be used. The literature along this line of work includes \cite{Zheng2016, Hutter2018, Wang2018, Cheng2020, Liu2020}. In these works, to make the line segment of the pointer salient, they first apply an adaptive threshold method to binarize the image. A thinning (also known as skeletonization) algorithm can then refine the pointer's thickness as one pixel wide. Finally, Hough-based line detection is used to detect all the straight lines in the image. 

However, despite being easy to implement, this branch of methods comes at a cost in generalizability:

\begin{enumerate}
\item[1)]The adaptive binarization is prone to lousy segmentation under extreme illumination conditions, which cannot tell the difference between the pointer and the background, inducing noise to the binary image.

\item[2)]The line representing the pointer could break apart during the process if it is too thin, inflicting difficulties in separating them with scales, captions, or borders of the reading area.

\item[3)]The lines detected by the Hough transform is directionless, meaning that the pipeline must integrate extra steps to distinguish the direction of the line, usually by locating the pointer's rotation center \cite{Zheng2016, Hutter2018, Wang2018}, or by projection to a specific axis of the image \cite{Gao2018}. However, not all kinds of meters have a visible spindle, and it is hard to correlate the line with the center without handcrafted rules. 
\end{enumerate}

By and large, without individual fine-tuning, determining the pointer's position and direction is a non-trivial task for these methods to generalize to different instruments in various scenarios.

Instead of leveraging the aforementioned image-processing techniques, some researchers turn to learning-based approaches to fully exploit a learned model's power in terms of universality and robustness to damping factors. Reference \cite{Yang2018} proposed a CNN-based scheme, wherein the pinpoint of the pointer is located as a landmark in the dial face with a key point detection idea like ours. Reference \cite{Lin2020} estimated the pointer binary mask with a network modified from the Mask-RCNN and then fitted the mask to get the pointer's centerline using the PCA algorithm. Reference \cite{Alexeev2020} employed a convolutional neural network with a non-linear Network in Network kernel, using global regression to estimate grids of reference points for all symbols (scales) on the meter panel and the position of the pinpoint. 

These methods have shown their superiority over traditional methods regarding accuracy and versatility in experiments yet require no parameter-sensitive preprocessing such as binarization or brightness histogram balancing. However, they did not simultaneously estimate the direction of the pointer but rely on extra features to orient it, which, as argued before, could be obscured or missing in certain meter types. Besides, the methods of \cite{Lin2020, Alexeev2020} cannot generalize to meters equipped with multiple pointers without a drastic revision of the network.

\section{Methodology}

In this section, we firstly describe the AMR pipeline in Sec. \ref{Sec:3-A}, which processes the images captured by a patrol robot to read off pointer-type analog meters in industrial inspection scenarios. Note that the pipeline is proposed to holistically clarify the functionality and data I/O of the pointer detection method as a part of it, but not persuade the readers to rigidly follow the methods (although it could be a good start) we used here, especially for meter detection and template matching, since they could harness state-of-the-art in the domain of image processing and artificial intelligence. On the contrary, we focus on presenting the Vector Detection Network (VDN) as our pointer detection method, describing its network architecture, training paradigm, and optimization in Sec. \ref{Sec:3-B}, Sec. \ref{Sec:3-C}, and Sec. \ref{Sec:3-D} respectively.

\subsection{Reading off Analog Meters using Robots}\label{Sec:3-A}

In industrial inspection scenarios, the patrol robot acts analogous to a human inspector, whose telescope and eyes are replaced by a zoom camera integrated into an actuated pan-tilt mounted atop the robot. Given the locations of the observing spots (known as checkpoint) in the scene and the viewing angle at each spot, the robot navigates itself from one spot to another, posing the pan-tilt at each checkpoint, and reading off analog meters following the pipeline (depicted in Fig. \ref{fig:2}) consisting of four steps:

\subsubsection{Meter Detection}
The meter panels are detected by a Region-based Convolutional Network (R-CNN) \cite{He2017} from the image $I$ shot by the robot's onboard camera, whose framing, focus, and exposure parameters are calibrated beforehand to make the meters crystal clear in the image. Optional fine adjustments could be leveraged afterward to center the panel of interest, search around if the default setting failed, or update the capturing parameters based on the detection quality. The bounding boxes of meters yield by R-CNN are used to crop $I$, producing image patches $\mathcal{P}^I=(\mathcal{I}_1,\mathcal{I}_2,\ldots,\mathcal{I}_m)$, where $m$ is the number of meters detected in $I$.

\subsubsection{Pointer detection}
The pointer(s) is/are retrieved from every image in the patch set $\mathcal{P}^I$ with the vector detection approach detailed in Sec. \ref{Sec:3-B}. The resulting vectors in each patch are leveraged in the numerical calculation as detailed below to derive the gauge's readings indicated by the patch. 

\subsubsection{Template Matching}
Let $T$ be a template image captured at the same spot as $I$ and manually labeled with $m^*$ (may not equal to $m$) bounding boxes for meters of interest; we generate template patches $\mathcal{P}^T=(\mathcal{T}_1,\mathcal{T}_2,\ldots,\mathcal{T}_{m^*})$ by cropping $T$ with the given bounding boxes. Our goal is to associate as many elements in $\mathcal{P}^T$ to those in $\mathcal{P}^I$ and establish the pixel-to-pixel correspondence between the matched pairs. To this end, we first assign the bounding boxes in $I$ to the labeled ones in $T$ via the Hungarian algorithm \cite{Kuhn1955}, pairing $\mathcal{P}^I$ with $\mathcal{P}^T$ by minimizing the overall geometric center distance for all matches. This coarse matching works because both $I$ and $T$ are captured with the same configuration (in this context, it means the same checkpoint, viewing angle, and camera parameters); hence the center of the same meter in both images should be adjacent to each other than to others. This trick outperforms directly applying global keypoint matching to the images $\langle I, T \rangle$ because finding stable keypoints and optimizing the assignment could be both susceptible to the background and time-consuming for these two high-resolution images. Nevertheless, for the matched pairs $\langle \mathcal{I}_i,\mathcal{T}_j \rangle$, we then leverage the keypoint-based matching algorithm \cite{Luo2019} to establish the pixel-to-pixel correspondence represented by the homography matrix $\mathbf{M}_{ji}$, since the matching areas are constrained to the foreground (meter panels). 

\subsubsection{Reading Calculation}
For each meter, with the homography matrix $\mathbf{M}_{ji}$ derived from the matched pair $\langle \mathcal{I}_i,\mathcal{T}_j \rangle$, where $i\in[1, m]$ and $j\in[1,m^*]$, we project the pixel coordinates $\mathbf{s}^{\mathcal{T}_j} \in \mathbb{R}^{e \times 2}$ ($e$ is the scale point number) of the gauge's scale points formerly labeled in the template $\mathcal{T}_j$ onto $\mathcal{I}_i$ to derive

\begin{equation}
\mathbf{s}^{\mathcal{I}_i}=\mathbf{M}_{ji} \mathbf{s}^{\mathcal{T}_j},
\label{eq:1}
\end{equation}
where $\mathbf{s}^{\mathcal{I}_i} \in \mathbb{R}^{e \times 2}$ is the projected scale point set in $\mathcal{I}_i$. Then we interpert each detected pointer vector in $\mathcal{I}_i$ as a ray, which should intersect with a line segment formed by consecutive scale points $(\mathbf{s}_q, \mathbf{s}_{q+1}) \in \mathbf{s}^{\mathcal{I}_i}, q \in [0, e-1]$, yielding an intersection (say, point $\mathbf{p}^{\mathcal{I}_i}_k \in \mathbb{R}^2$ for the $k$-th detected vector, see Fig. \ref{fig:2}(e) for a visual illustration). Let $v_q$ and $v_{q+1}$ be the scale values of $\mathbf{s}_q$ and $\mathbf{s}_{q+1}$ respectively, the $k$-th pointer's numeric reading $r_k$ is

\begin{equation}
r_k=\frac{\left\|\mathbf{s}_q \mathbf{p}^{\mathcal{I}_i}_k\right\|}{\left\|\mathbf{s}_q \mathbf{s}_{q+1}\right\|}(v_{q+1}-v_q) + v_q.
\label{eq:2}
\end{equation}

Notice that the final readings are relevant to the semantics of each pointer and the unit. Combining all numeric readings to one or multiple meaningful readings requires a case by case handling. In practice, for meters having two or more independent dials as the upper left inset of Fig. \ref{fig:1}(b), it is suitable to exploit the k-nearest-neighbors (KNN) algorithm to assign each pointer to the corresponding dial plate as long as the positions of all pinpoints and scale points are derived. Hence \ref{eq:2} could be applied individually. For pointers sharing the same spindle and dial face like those in Fig. \ref{fig:2}, prior knowledge of their relative positions could be exploited to develop such a strategy. 

\begin{figure*}[t]
\centering
\vspace{6pt}
\includegraphics[width=7.16in]{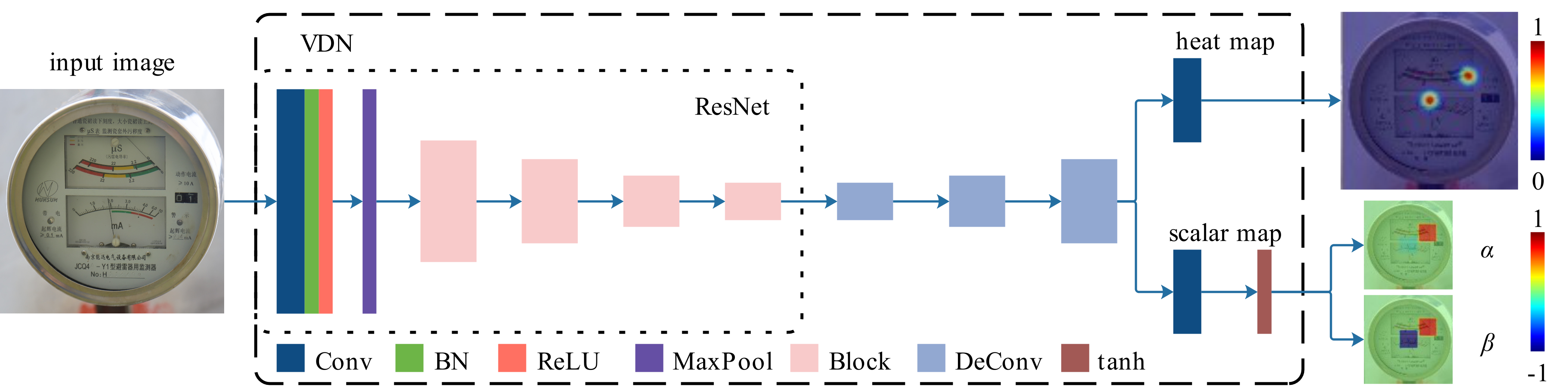}
\caption{The architecture of VDN. The input image is first processed by a conventional \emph{ResNet}, whose configurations are as detailed in \cite{He2016}. Then, three deconvolution layers increase the spatial dimension of the extracted features. Finally, two branches respectively estimate a confidence heat map and a two-layer scalar map via convolution. Note that we apply \emph{tanh} to the scalar maps to make the output values within the $[-1,1]$ range.}
\label{fig:3}
\end{figure*}

\subsection{Vector Detection Network}\label{Sec:3-B}

Given a set $\mathcal{P}^I$ of image patches, we iteratively take one patch $\mathcal{I}_i \in \mathcal{P}^I, i\in[1,m]$ and resize $\mathcal{I}_i$ to $\mathcal{I}_i'$ with an affine transform. Accordingly, the proposed Vector Detection Network (VDN) takes, as input, $\mathcal{I}_i'$ of size $w' \times h'$, and produces, as output, the tensors representing estimated unit vectors along with their confidence coefficients as being pointers. Formally, we target to obtain the set $\hat{\Delta}=(\hat{\delta}_1,\hat{\delta}_2,\cdots,\hat{\delta}_{\hat{\eta}})$ via VDN. Let

\begin{equation} \label{eq:3}
\hat{\delta}_k=\{\hat{x}_k, \hat{y}_k, \hat{\alpha}_k, \hat{\beta}_k, \hat{c}_k\}
\end{equation}
be the detection result of the $k$-th vector, where $(\hat{x}_k, \hat{y}_k)$ is the initial point; $\hat{\alpha}_k \in[-1,1], \hat{\beta}_k \in[-1,1]$ are the scalar components on $x$-axis and $y$-axis respectively; $\hat{c}_k$ is the confidence coefficient; $k\in[1, \hat{\eta}]$. $\hat{\eta}$ is the number of detected pointers. Thus VDN ($f$) could be denoted as

\begin{equation} \label{eq:4}
\hat{\Delta}=f(\mathcal{I}'_c).
\end{equation}

Note that the actual pointer number ($\eta$) of a particular meter could be known beforehand, yet VDN itself is agnostic to $\eta$, and only outputs confirmed pointers regarding the confidences, such that $\hat\eta$ may not equal to $\eta$ due to inadequate confidence criteria or pointers being overlapped with each other.

To formulate $f$, we referred to the human keypoint detection techniques and let $f$ be analogue to the human pose estimation network proposed in \cite{Xiao2018}. We assume that if such a method holds for limb joints, it could also be adopted to points with stable semantics, such as the spindle or the tip of a pointer. We verified the hypothesis and found the superiority of leveraging pinpoints as the target. 

Accordingly, VDN illustrated in Fig. \ref{fig:3} cascades a feature extraction backbone network (ResNet \cite{He2016}), three deconvolutional layers, and a $1\times1$ convolutional layer for predicting a 2D confidence map $\mathrm{\hat{H}}$ in the form of a heatmap. However, VDN distinct from \cite{Xiao2018} in two aspects. 

First, \cite{Xiao2018} only outputs heatmaps, whereas VDN outputs a two-layer $\textit{scalarmap}$ denoted as $\hat{\mathrm{V}}$ via an extra convolutional layer aside one heatmap $\mathrm{\hat{H}}$. Both $\mathrm{\hat{H}}$ and $\mathrm{\hat{V}}$ are of size $w^*\times h^*$ in terms of the last two dimensions, where $w^*=\lambda w', h^*=\lambda h'$, and $\lambda<1$ is a scale factor determined by the network architecture. Second, in \cite{Xiao2018}, the yield heatmap tensor has multiple channels, one per joint, which jointly assemble one instance (person in this case), and since each heatmap channel takes care of one limb joint, the individual human keypoint is the global maximum of that channel. Whereas VDN outputs a heatmap with only one channel because $\eta$ is unknown beforehand. We turn to local maximums to locate an arbitrary number of peaks. By doing this, we can simultaneously derive multiple pointers, avoiding complex grouping operations applied to human limb joints \cite{Cao2017}.

In Fig. \ref{fig:3} we also depict how $\mathrm{\hat{H}}$ and $\mathrm{\hat{V}}$ convey $\hat{\delta}_k$. A peak in $\mathrm{\hat{H}}\in \mathbb{R}^{w^* \times h^*}$ is consided as a pointer tip if its pixel value $\hat{c}_k$ exceeds the confidence threshold. If a peak meets the criterion, its coordinates $(\hat{x}^*_k, \hat{y}^*_k)$ in $\mathrm{\hat{H}}$ could be scaled back to $\mathcal{I}_i'$, producing $(\hat{x}_k, \hat{y}_k)$ with $\lambda$. As for $\hat{\alpha}_k$ and $\hat{\beta}_k$, we query each channel of $\mathrm{\hat{V}}$ at the location $(\hat{x}^*_k, \hat{y}^*_k)$, and assign the two values to $\hat{\alpha}_k$ and $\hat{\beta}_k$ respectively.

\subsection{Training Objective}\label{Sec:3-C}

To activate a data-driven approach like VDN and facilitate research on the AMR domain, we established a dataset consisting of pointer-type analog meters captured in real scenarios. This effort is essential due to few public datasets of this kind are available to date. Such a dataset promotes deep learning models like VDN, which needs to be trained with diverse and abundant data to be generalizable. We also hope that it could enhance the researchers' understanding of this problem since some of the samples are rare to occur or hard to obtain. A detailed introduction of this dataset named Pointer-10K is presented in Sec. \ref{Sec:4}.

To construct the training objective with the dataset, for each sample image $S$ in the train split, we first leverage corresponding labeled bounding boxes to crop the image and produce the sample patch set $\mathcal{P}^S=(\mathcal{S}_1,\mathcal{S}_2,\cdots,\mathcal{S}_n)$. After applying data argumentations, namely random scaling and rotation to $\mathcal{S}_i, i \in [1,n]$, where $n$ is the meter number in $S$, we transform $\mathcal{S}_i$ to $\mathcal{S}' _i$ with an affine transform to yield the input for VDN. Then, following the same fashion as \eqref{eq:3}, we let $\Delta$ be the groundtruth reference for VDN processing $\mathcal{S}'_i$, where $\delta_k=(x_k, y_k, \alpha_k, \beta_k, 1) \in \Delta, k\in[1, \eta]$. To derive the groundtruth heatmap $\mathrm{H}$ and scalarmap $\mathrm{V}$ with $\Delta$, we leverage two pixels manually labeled for every pointer in $\mathcal{S}_i$---one located on the pointer's tail and the other on the tip by the following steps. 

First, to generate the groundtruth heatmap corresponding to $\mathcal{S}' _i$, we consider the pointer tip pixel $\mathbf{p}_k=(x_k, y_k)$ as the position where the initial point of the vector estimated by VDN should regress to. This choice is made because, in most real-world scenarios, the tip of the pointer is sharper and easier to locate, whereas other parts of the pointer, such as the spindle or the tail, could be bold or invisible, making them hard to label. In this sense, the value of each pixel on $\mathrm{H}\in \mathbb{R}^{w^* \times h^*}$ represents the belief that a pinpoint locates at that pixel, for each $\mathbf{p}_k$, we scale it to $\mathbf{p}^*_k=(\lambda x_k,\lambda y_k)$, and calculate $\mathrm{H}$ by applying a 2D Gaussian centered on $\mathbf{p}^*_k$

\begin{equation}\label{eq:5}
\mathrm{H}(\mathbf{\rho})=\mathrm{exp}\left(-\frac{\min_{k=1}^{\eta}(\|\mathbf{\rho}-\mathbf{p}^*_k\|^2_2)}{2\sigma^2}\right),
\end{equation}
where $\mathbf{\rho}$ is an arbitrary pixel on $\mathrm{H}$; $\sigma$ is the standard deviation. 

Second, to generate the groundtruth scalarmap for $\mathcal{S}' _i$, we recall that by its definition, the scalarmap $\mathrm{V} \in \mathbb{R}^{w^*\times h^*\times 2}$ encodes the pointing directions of pointers with its two channels $\mathrm{V}_\alpha$ and $\mathrm{V}_\beta$ respectively, where the values $\alpha$ and $\beta$ represent the two scalar components of a vector. We initialized $\mathrm{V}_\alpha$ and $\mathrm{V}_\beta$ with zeros and accumulated the value of each pixel $\mathbf{\varrho}$ of them $\forall k\in[1,\eta]$ in a recursive manner. Let

\begin{align}\label{eq:6}
\mathrm{V}_{\alpha,k}(\mathbf{\varrho})=
\begin{cases}
\alpha_k & \text{if $\|\mathbf{\varrho}-\mathbf{p}^*_k\|\leq3\sigma$,} \\
0 & \text{elsewise,}
\end{cases}
\\
\mathrm{V}_{\beta,k}(\mathbf{\varrho})=
\begin{cases}
\beta_k & \text{if $\|\mathbf{\varrho}-\mathbf{p}^*_k\|\leq3\sigma$,} \\
0 & \text{elsewise,}
\end{cases}
\end{align}
where $\mathrm{V}_{\alpha,k}(\mathbf{\varrho})$ and $\mathrm{V}_{\beta,k}(\mathbf{\varrho})$ respectively denotes the $k$-th pointer's contributed value to the two channels of $\mathrm{V}$ at pixel $\mathbf{\varrho}$. Then we have

\begin{align}\label{eq:8}
\mathrm{V}_\xi(\mathbf{\varrho})=
\begin{cases}
\sum\limits_{k=1}^{\eta}\mathrm{V}_{\xi, k}(\mathbf{\varrho}) / \mathrm{C}(\mathbf{\varrho}) & \text{if $\mathrm{C}(\mathbf{\varrho})>0$,} \\
0 & \text{elsewise,}
\end{cases}
\end{align}
where $\xi \in [\alpha, \beta]$. $\mathrm{C}\in\mathbb{R}^{w^* \times h^*}$ is a cache tensor initialized by zeros, and takes the form 

\begin{align}\label{eq:9}
\mathrm{C}(\mathbf{\varrho})=\sum\limits_{k=1}^{\eta}\Psi({\mathrm{V}_{\xi,k}}), \text{ where }
\Psi(x)=
\begin{cases}
1 & \text{if $x > 0$} \\
0 & \text{elsewise,}
\end{cases}
\end{align}
With \eqref{eq:8} and \eqref{eq:9}, $\mathrm{V}$ was derived by stacking $\mathrm{V}_\alpha$ and $\mathrm{V}_\beta$ along the channel dimension. 

\subsection{Optimization}\label{Sec:3-D}

The training of VDN aims to minimize the Mean Squared Error (MSE) loss between the predicted heatmap $\hat{\mathrm{H}}$ and the target heatmap $\mathrm{H}$ denoted as $l_{(\hat{\mathrm{H}}, \mathrm{H})}$, along with the MSE loss $l_{(\hat{\mathrm{V}},\mathrm{V})}$ between $\hat{\mathrm{V}}$ and $\mathrm{V}$. The overall lose $l$ is

\begin{equation}\label{eq:10}
l(\epsilon)=l_{(\hat{\mathrm{H}},\mathrm{H})}+\mu\frac{\epsilon}{E}l_{(\hat{\mathrm{V}},\mathrm{V})},
\end{equation}
where $E$ is the total training epochs, $\epsilon\in[0,E-1]$ is the epoch index, and $\mu$ is a constant impeding $l_{(\hat{\mathrm{V}},\mathrm{V})}$ to dominate the early stage of training. The loss $l_{(\hat{\mathrm{V}},\mathrm{V})}$ is specifically tackled as this because it converges slower than $l_{(\hat{\mathrm{H}},\mathrm{H})}$, making VDN be prone to stuck in a local maximum. From an empirical view, the geometric context around the pointer tip strongly enphasises the peak in $\mathrm{\hat{H}}$, but for $\mathrm{\hat{V}}$ the context is less significant and only by a period of exploring could the model establish a steady estimation of $\mathrm{\hat{V}}$. 

\begin{figure*}[t]
\centering
\vspace{6pt}
\includegraphics[width=7in]{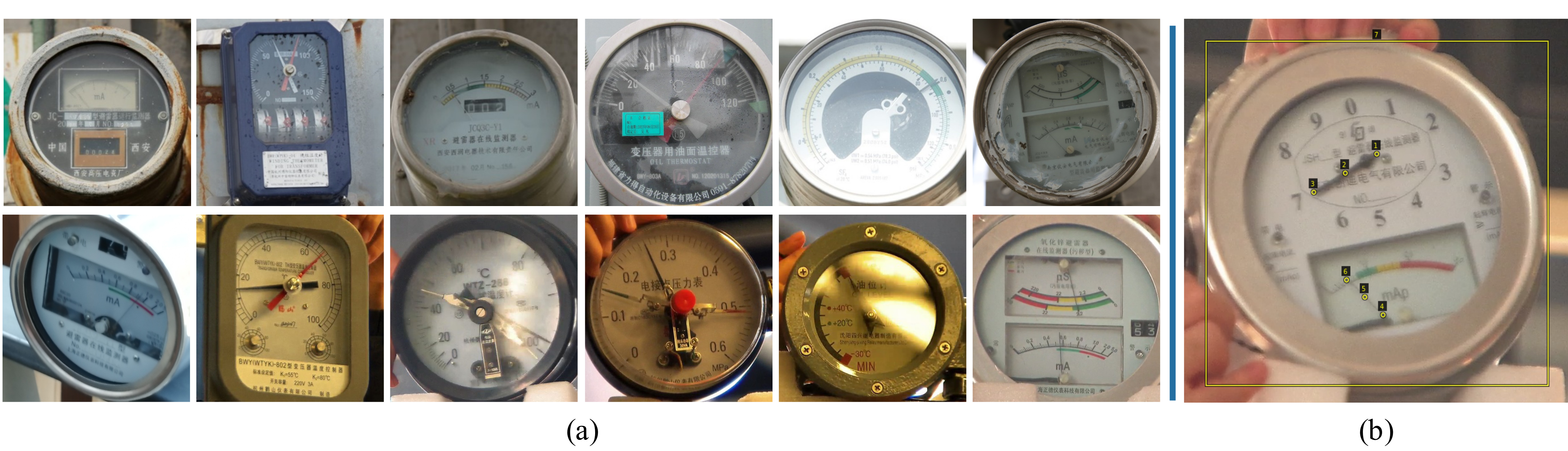}
\caption{A non-exhaustive illustration of the samples from the Pointer-10K dataset. We removed the background to save space and make the pointer more obvious. (a) \textit{top row}: Samples from real scenarios; \textit{bottom row}: Synthetic samples of offline meters whose pointers were manually posed. (b) Annotation of the sample image labeled with the VGG Image Annotator (VIA) \cite{Dutta2019}.}
\label{fig:4}
\end{figure*}

\section{Dataset}\label{Sec:4}

We present the Pointer-10K dataset containing 5440 images and approximately 10,000 pointer instances as a significant contribution to the research community. A non-exhaustive illustration of the meters in the dataset is shown in Fig. \ref{fig:4}. The main challenge for such a dataset is to preserve the complexity of real-world scenarios whereas being unbiased to different meter types, readings, or environments. To tackle this, we composed the dataset with images captured in real and synthetic situations with those obtained online. 

For the real portion, since the dataset is originated from a project leveraging a wheeled patrol robot to aid integrated substation automation, the images are mainly for analog meters wildly adopted in substations, including arrester monitors, gas manometers, and thermometers. The images were either captured on-site with the robot's onboard pan-tilt camera or captured manually with cellphones or a digital single-lens reflex (DSLR). During the capturing, the meters' readings were not able to be manipulated. Hence in most cases, the readings were constrained in normal ranges, which is not beneficial for the training of an unbiased pointer detection network (and thus the synthetic data are needed). However, these real data provide valuable insight into what the actual industrial environment looks like. Besides, the viewpoints only constrained by the visibility of the meter were chosen randomly to diversify the data. The viewing distance ranged from 1m to 10m (with zooming) approximately, and the viewing angle was not restricted to be perpendicular to the dial plate but only guaranteed that the pointers and scales could be seen. The real portion makes up 1/3 of the total samples.

For the synthetic portion, please note that by \emph{synthetic} we mean manually rotating the pointers to add more rare cases to the dataset (e.g., reading out of the normal range or even the measurement range). Nevertheless, we do not claim that we have covered every possible reading through such procedure, nor is that necessary. In contrast, we consider the synthesis a key factor for obtaining adequate data to strengthen the dataset and methods raised from it. We show by the implementation that the proposed model generalizes well to arbitrary readings given finite training samples despite the significant change in surroundings, implying varying the poses of pointers is crucial for a learn-based method such as VDN. For this portion of data taking 1/3 of all samples, we captured available meters brought by us with the same devices mentioned before in both indoor and outdoor scenarios. The viewing standard was kept the same as for real data.

For the online portion, we made a crawler program to grasp industrial gauge images from the Internet. This portion makes up 1/3 of all data.

We did not provide a per-category label for each kind of meter since tackling them case by case is what we intend to avoid. Instead, we labeled all meters with the same label, i.e., \emph{meter}, which is sufficient for training a generic meter detection network. The dataset was labeled with the VGG Image Annotator \cite{Dutta2019} and then converted to the COCO format \cite{Lin2014}, offering bounding boxes of meters and three key points for each pointer---the tip, the tail, and the midpoint automatically calculated in between as depicted in Fig. \ref{fig:4}(b). Note that when the tail is not visible, we instead label the closet point to the tail that could be seen.

We split the dataset in a $7:2:1$ fashion for training (\emph{train-pointer}), validation (\emph{val-pointer}), and testing (\emph{test-pointer}), respectively. We manually guaranteed that the test split contains 40\% hard samples, whose types do not exist in the other two sets. 

The imaging quality was not constrained, and for some samples, we even deliberately blurred them or worsened the illumination. Experiments suggest that such handling is constructive for VDN to deal with real-world scenarios and avoid a bias to specific imaging devices. For images with a raw resolution larger than $1280\times720$ pixels, we scaled them down to 1280 pixels width while kept the aspect ratio unchanged. Whereas for smaller images, we just kept them intact to prevent distortion.

\section{Experiments}

\begin{table*}[t]
\centering
\vspace{6pt}
\begin{threeparttable}
\caption{Evaluation on Pointer-10K Dataset (See the main text in Sec. \ref{Sec:5-B} for details)} \label{tab:1}
\begin{tabular}{lcccccccccccc} 
	\toprule
	{} & {Split} & {Metric} & {$\mathrm{AP}$} & {$\mathrm{AP}_{50}$} & {$\mathrm{AP}_{75}$} & {$\mathrm{AP}_{M}$} & {$\mathrm{AP}_{L}$} & {$\mathrm{AR}$} & {$\mathrm{AR}_{50}$} & {$\mathrm{AR}_{75}$} & {$\mathrm{AR}_{M}$} & {$\mathrm{AR}_{L}$}	\\
	\midrule
	\multirow{4}{*}{VDN\_ResNet18\_384x384} & \multirow{2}{*}{\emph{val-pointer}} & OKS & 0.972 & 0.980 & 0.970 & 0.984 & 0.973 & 0.978 & 0.981 & 0.979 & 0.986 & 0.979 \\
	{} & {} & VDS & 0.851 & 0.899 & 0.864 & 0.945 & 0.980 & 0.904 & 0.932 & 0.912 & 0.948 & 0.984 \\
	{} & \multirow{2}{*}{\emph{test-pointer}} & OKS & 0.953 & 0.958 & 0.958 & 0.972 & 0.953 & 0.960 & 0.965 & 0.964 & 0.979 & 0.959 \\
	{} & {} & VDS & 0.776 & 0.832 & 0.799 & 0.913 & 0.946 & 0.859 & 0.893 & 0.872 & 0.919 & 0.948 \\
	\bottomrule
\end{tabular}
\end{threeparttable}
\end{table*}

\subsection{Implementation Details} \label{Sec:5-A}

We implemented VDN using PyTorch. The \emph{train-pointer} split was leveraged for training while \emph{val-pointer} and \emph{test-pointer} were used for performance tests and ablation tests. Note that \emph{val-pointer} was also utilized for determining the hyperparameters, including the batch size, the learning rate schedule, the epoch number, and $\mu$ in \eqref{eq:9}. The training and evaluations were conducted on a desktop with an AMD 3700X CPU, an NVIDIA GeForce 2070 GPU, and 16GB RAM.

The size of input images was set as $384\times 384$ pixels. Unless otherwise noted, the backbone network was a ResNet18 initialized by the model pre-trained on the ImageNet classification task \cite{Deng2009}. During training, the base learning rate was 1e-3 initially and then dropped to 1e-4 and 1e-5 at the 140th and 190th epoch, respectively. The 200 epochs in total took approximately 5 hours for training with eight samples per batch. $\mu$ in \eqref{eq:9} was set as 1 and $\sigma$ in \eqref{eq:5} was set as 3. We kept other hyperparameters as the same as \cite{Xiao2018}.

\subsection{Main Results} \label{Sec:5-B}

As VDN is designed to detect features that could be semantically described as vectors in an image, from a generic point of view, we examined its performance regarding the accuracy for initial points detection and orientation (scalar components) estimation as separately reported below. The result of the default configuration (ResNet18 backbone, $384 \times 384$ input size) was given in Table \ref{tab:1}. 

\subsubsection{Initial Point Detection}
We leveraged the standard object keypoint similarity (OKS) \cite{Coco2021} as the metric for this evaluation, which is calculated as

\begin{equation} \label{eq:11}
OKS=\frac{\sum_k[\exp(\frac{-d^2_k}{2\sigma^2_d})\varphi(\vartheta_k>0)]}
{\sum_k[\varphi(\vartheta_k>0)]},
\end{equation}
where $d_k$ is the Euclidean distance between the $k$-th detected initial point and its corresponding groundtruth; $\vartheta_k$ is the visibility flag of the groundtruth. Note that in our dataset all target keypoints are visible and labeled, such that $\varphi(\vartheta_k>0)\equiv 1 \quad \forall k \in [1,N]$; $N$ is the total number of pointer instances in the hold-out split. $\sigma_d=a_k\tau_k$ is the standard deviation of this gaussian, where $a_k$ is the object scale and $\tau_k$ is a per-keypont constant that controls falloff. Intuitively, $\sigma_d$ balances the numerical value of $d_k$ by letting $a_k$ be the area of the bounding box surrounding the dial plate where the $k$-th keypoint belong to, such that zooming the image will not influence OKS. Unless otherwise specified, we let $\tau_k=0.1$ in this work as all keypoints are of the same type and with the same importance.

\subsubsection{Direction Estimation}
We leveraged a tailored metric named Vector Direction Similarity (VDS) for evaluating the similarity between the estimated vector direction with the groundtruth. The included angle $\theta_k \in [0, 180^{\circ}]$ formed by the detected vector and the groundtruth is analogous to the distance $d_k$ in the OKS metric. Let

\begin{equation}
VDS=\frac{\sum_k[\exp(\frac{-\theta^2_k}{2{\sigma^2_{\theta}}})\varphi(\vartheta_k>0)]}{\sum_k[\varphi(\vartheta_k>0)]},
\end{equation}
where $\varphi(\vartheta_k>0)$ is as explained in \eqref{eq:10}. We set $\sigma_\theta=a_k\kappa_k$, and $\kappa_k=0.2$ in the experiments. 

Both OKS and VDS were implemented with \emph{cocoapi}. The standard metrics, including $\mathrm{AP}$ (average precision), $\mathrm{AP}_{50}$, $\mathrm{AP}_{75}$, $\mathrm{AP}_M$, and $\mathrm{AP}_L$ were used for characterizing the precision of VDN in terms of OKS and VDS. Meanwhile, AR (averaged recall), $\mathrm{AR}_{50}$, $\mathrm{AR}_{75}$, $\mathrm{AR}_M$, and $\mathrm{AR}_L$ were used to measure the sensitivity of VDN. For a more in-depth explanation of these metrics, see \cite{Coco2021a}.

The main results indicate that our method generalizes to unseen samples without a huge deterioration regarding accuracy or sensitivity (1.9\% drop on percision and 1.8\% on recall) even there are 40\% hard examples in the test split. The results also show that pointers occupying larger areas in the dial plate are more likely to be located and orientated correctly, given that $\mathrm{AP}_M < \mathrm{AP}_L$ and $\mathrm{AR}_M < \mathrm{AR}_L$ for both OKS and VDS.

We emphasize that an accurate pointer detection does not naturally lead to an accurate reading of the meter since that also depends on other modules of the AMR pipeline. With that said, performing a comprehensive comparison of the reading accuracy with conventional works is out of our scope. Nevertheless, one can implement the rest of the pipeline or benchmark VDN based on our publicly available code at \url{https://github.com/DrawZeroPoint/VectorDetectionNetwork}. Following the scheme described in Sec. \ref{Sec:3-A}, we were able to archive an error rate less than $\pm3\%$ in a real-world scenario.

\begin{table}[t]
\centering
\vspace{6pt}
\begin{threeparttable}
\caption{Backbone Network Comparision on test-pointer} \label{tab:2}
\begin{tabular}{lcccc}
	\toprule
	Backbone  & $\mathrm{AP}^\mathrm{kp}$ & $\mathrm{AP}^\mathrm{vd}$ & Rate (fps) & Size (MB) \\
	\midrule
	ResNet18  & 0.953 & 0.776 & 171 & 59 \\
	ResNet34  & 0.953 & 0.775 & 119 & 97 \\
	ResNet50  & 0.942 & 0.768 & 81 & 129 \\
	\midrule
	Res2Net50 & 0.953 & 0.783 & 67 & 130\\
	\bottomrule
\end{tabular}
\end{threeparttable}
\end{table}

\subsection{Ablation Experiments} \label{Sec:5-C}

We report the performance of VDN under two kinds of ablations, namely, backbone architecture and input variation.

\subsubsection{Backbone Architecture}

For this experiment, we altered the backbone of VDN while kept other configurations unchanged. As shown in Table \ref{tab:2}, at least on the proposed test set, deepen the ResNet backbone network of VDN could not significantly increase either OKS or VDS. Although it could be benefited from sophisticated design such as \cite{Gao2019}, such improvement also means a trade-off between accuracy and computational/memory overheads. In practice, we consider the default setting with ResNet18 to be decent in pointer detection scenarios. 

\subsubsection{Input Variation}

A common question for a neural network approach like VDN could be raised regarding their superior performance---whether the performance indicates a semantic cognition of the image context or just an over-fitting of the dataset? To address this question, we developed two experiments that drastically modified the input, such that the model could not excel unless catching the vital information for detecting the pointers.

\begin{figure}[t]
\centering
\includegraphics[width=3.5in]{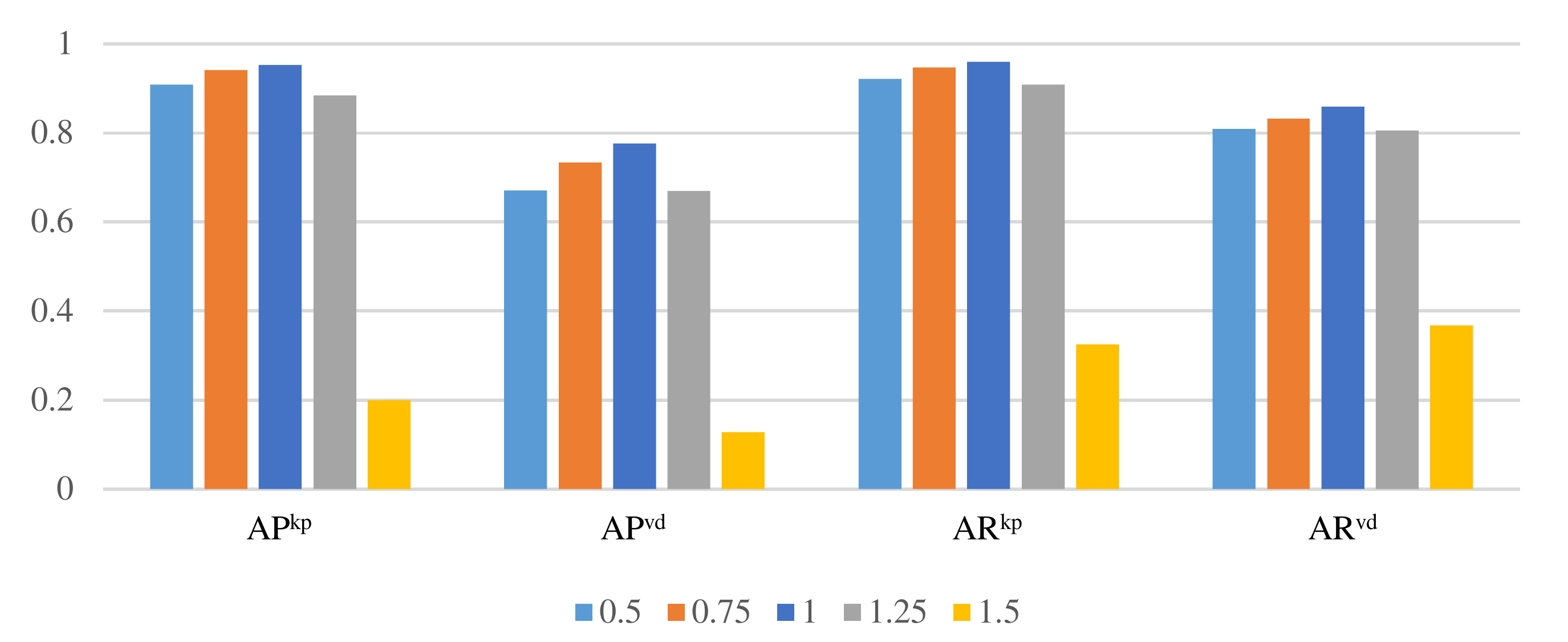}
\caption{OKS and VDS results regarding different scaling factors. The superscript kp is for OKS metric and vd for VDS metric. See the main text for detailed description.}
\label{fig:5}
\end{figure}

First, we scaled the input patch while kept the input dimensions the same as $384\times384$, such that a scaling factor larger than one means cropping the borders; otherwise, the borders are padded with zeros. Beyond the original scaling factor $s\in[0.98, 1.02]$ applied during training, we respectively adopted 0.5, 0.75, 1.25, and 1.5 for scaling. The results depicted in Fig. \ref{fig:5} show that both OKS and VDS decline when manipulating the scale. However, the performance only dropped drastically when cropping and enlarging the input image by a large margin ($s=1.5$), which is reasonable since in this case, the pointer's tip may be out of the visible zone. Yet for other cases, VDN still proved its generalization given that patches with $s=[0.5,0.75,1.25]$ were not trained beforehand. We empirically ascribe this to the ability of VDN to catch adequate contexts for the detection, including the pointers, the dial-plate, and their relationship, in the condition that they survive shrink or enlargement operations.

\begin{figure}[t]
\centering
\includegraphics[width=3.5in]{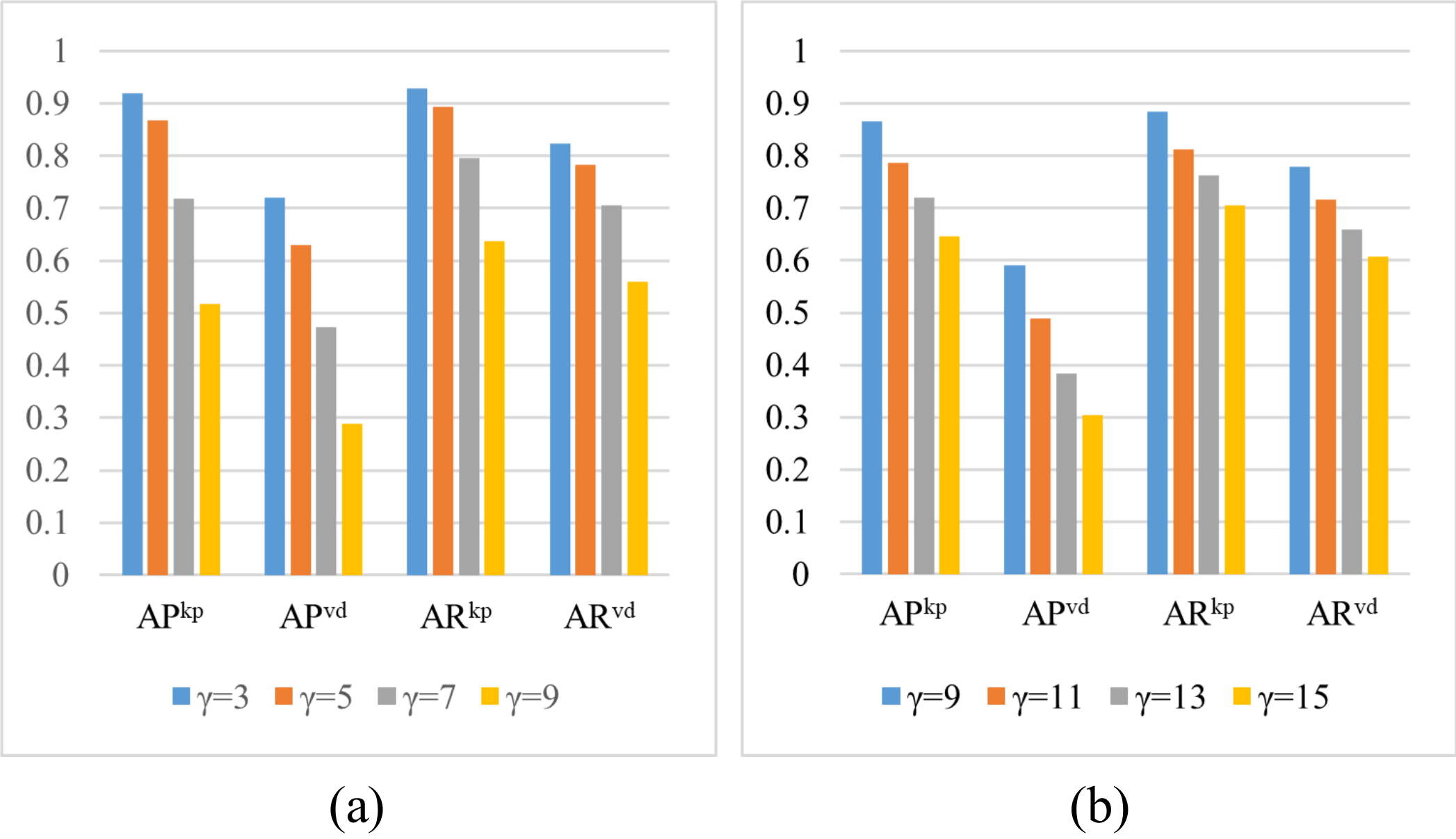}
\caption{OKS and VDS results yield by masking a square area centered at (a) the tip of the pointer and (b) the tail of the pointer under different mask sizes.}
\label{fig:6}
\end{figure}

\begin{figure*}[t]
	\centering
	\vspace{6pt}
	\includegraphics[width=7.16in]{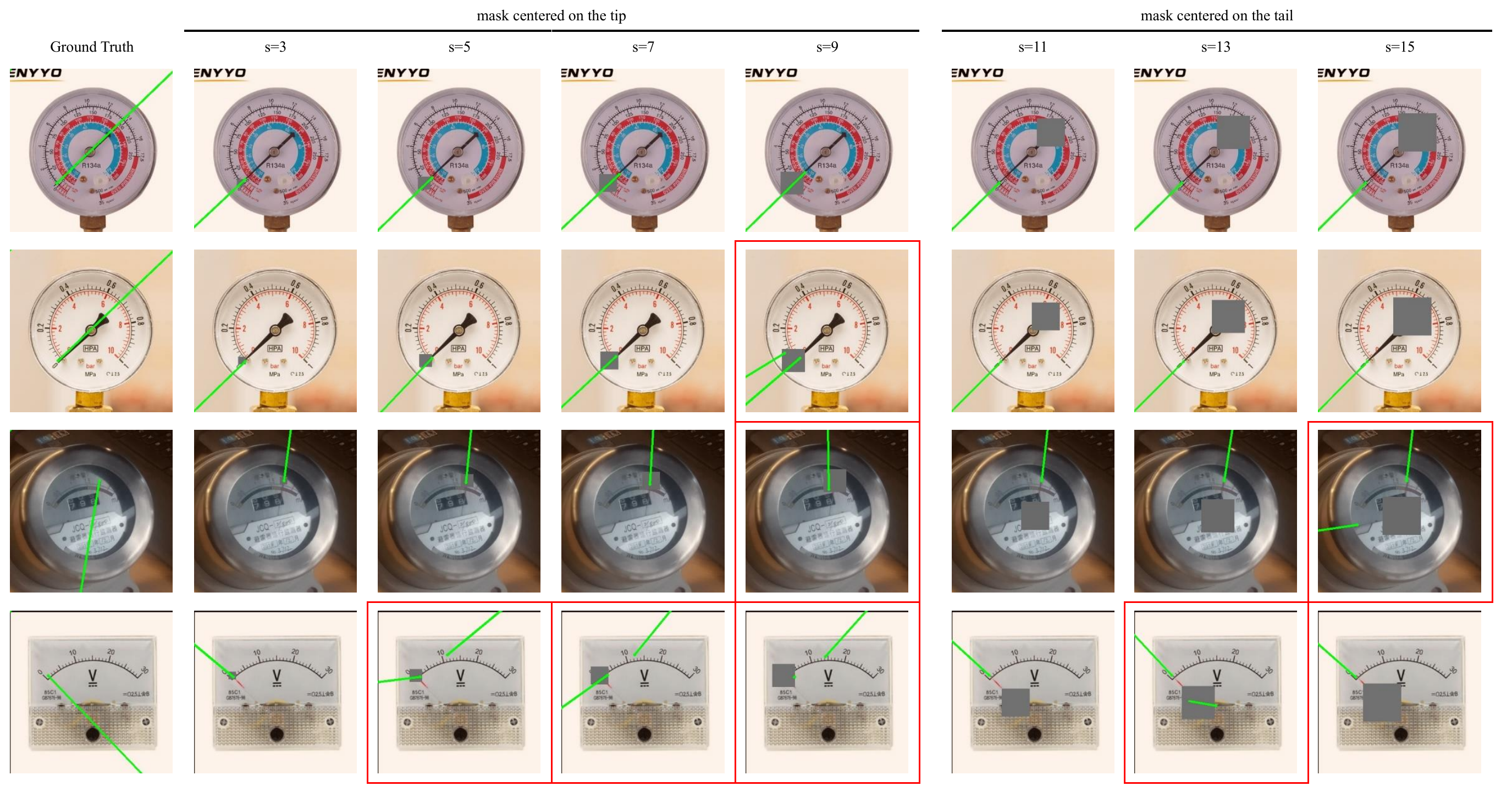}
	\caption{Qualitative results of VDN yield by applying masks (grey square centered on the tip/tail of the pointer, best viewed in color) to the input patches. First column: the groundtruth; Middle four columns: mask of different widths centered on the tip; Rightmost three columns: masks centered on the tail. The green rays represent the detected vectors. The ground truths were reversely marked to be obvious. Failure cases were surrounded with red boxes, where typical failure types include false positive (e.g., 2nd row, $s=9$), wrong direction (e.g., last row, $s=5$), and initial point drift (3rd row, $s=9$).}
	\label{fig:7}
\end{figure*}

Second, within the input patches $\mathcal{P}^I $, we masked a square area by setting values within as zeros of size $\gamma\sigma+1$, whose center is located either on the tail or the tip of the pointer. For masks centered at tips we have $\gamma=[3, 5, 7, 9]$, while for those at tails $\gamma=[9, 11, 13, 15]$. The quantitative results depicted in Fig. \ref{fig:6} indicated that VDN could resist the information loss introduced by applying the mask to the input, and those less relevant or influential to the pointer tips caused a milder drop in terms of both OKS and VDS. Random qualitative results and failure cases are shown in Fig. \ref{fig:7}.

Finally, it is worth noting that VDN is robust to the rotation of the input patch since at the data pre-processing phase before training, we applied random rotation to the input. Specifically, the rotation range is $\pm90$ degrees. Hence, we performed no ablation test on that.

\section{Conclusions}

In this work, we addressed the detection of pointers in analog meter images captured in the wild. A novel human keypoint detection inspired neural network was proposed to tackle the pointer detection problem, which is a primary bottleneck in the meter reading pipeline. A new dataset consisting of approximately 10K pointer samples was contributed together with the method to boost the research on the pointer detection subject. The results from elaborate experiments suggested that the proposed method can detect multiple pointers from different meters without bells and whistles and run in real-time. In this sense, it is applicable to patrol robots inspecting industrial sites.

\begin{figure}[t]
	\centering
	\includegraphics[width=3.5in]{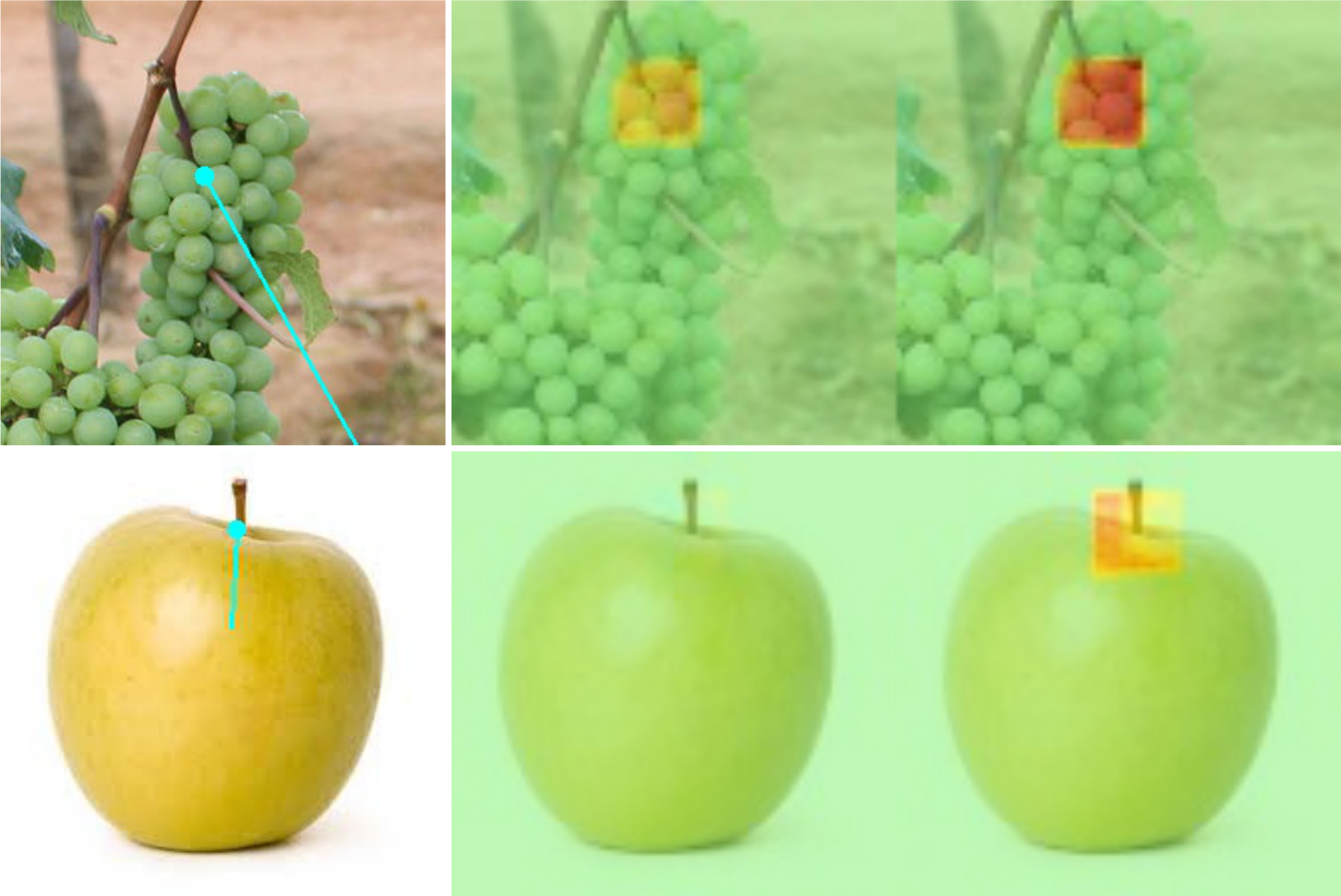}
	\caption{The implementation results of VDN trained on a dataset for detecting grape pedicels. Top row: the detected pedicel and the corresponding scalarmap indicating the keypoint's location and orientation. The color in the scalarmap follows the definition in Fig. \ref{fig:3}. Bottom row: VDN tailored for grape pedicels could detect that for an apple without fine-tuning.}
	\label{fig:8}
\end{figure}

In the future, we are about to apply VDN to other fields requiring vector detection in the wild. Such applications care about both the fine-grained line segments within images and also their directions. Potential scenarios include detecting stems or vines of plants, which could be crucial for agricultural robots aiding precise harvesting and automatic trimming. Some preliminary results are depicted in Fig. \ref{fig:8}, where we established a dataset for detecting pedicels of grapes. The only changed part was the dataset, whereas no modification was applied to VDN. Surprisingly, besides mastering the detection of pedicels, the trained model could yield reasonable output on other kinds of fruits, which do not belong to the training repertoire, indicating that the model is essentially sensitive to vector-like features.

\bibliographystyle{IEEEtran}
\bibliography{vdn}

\end{document}